\documentclass[runningheads]{llncs}
\usepackage{graphicx}
\usepackage{booktabs}
\usepackage{rotating}
\PassOptionsToPackage{hyphens}{url}\usepackage{hyperref}
\begin{document}
\title{IICONGRAPH: improved Iconographic and Iconological Statements in Knowledge Graphs}
\author{Bruno Sartini\inst{1}\orcidID{0000-0002-9152-4402}}
\authorrunning{B. Sartini}
\titlerunning{IICONGRAPH}
\institute{Ludwig-Maximilians-Universität München
\email{b.sartini@lmu.de}\\}
\maketitle              % typeset the header of the contribution
\begin{abstract}
Iconography and iconology are fundamental domains when it comes to understanding artifacts of cultural heritage. Iconography deals with the study and interpretation of visual elements depicted in artifacts and their symbolism, while iconology delves deeper, exploring the underlying cultural and historical meanings. Despite the advances in representing cultural heritage with Linked Open Data (LOD), recent studies show persistent gaps in the representation of iconographic and iconological statements in current knowledge graphs (KGs). To address them, this paper presents IICONGRAPH, a KG that was created by refining and extending the iconographic and iconological statements of ArCo (the Italian KG of cultural heritage) and Wikidata. The development of IICONGRAPH was also driven by a series of requirements emerging from research case studies that were unattainable in the non-reengineered versions of the KGs. The evaluation results demonstrate that IICONGRAPH not only outperforms ArCo and Wikidata through domain-specific assessments from the literature but also serves as a robust platform for addressing the formulated research questions. IICONGRAPH is released and documented in accordance with the FAIR principles to guarantee the resource's reusability. The algorithms used to create it and assess the research questions have also been made available to ensure transparency and reproducibility. While future work focuses on ingesting more data into the KG, and on implementing it as a backbone of LLM-based question answering systems, the current version of IICONGRAPH still emerges as a valuable asset, contributing to the evolving landscape of cultural heritage representation within Knowledge Graphs, the Semantic Web, and beyond.

\end{abstract}
\textbf{Resource type}: Knowledge Graph\\
\textbf{License}: Creative Commons Attribution 4.0 International\\
\textbf{DOI}: \url{https://doi.org/10.5281/zenodo.10294589}\\
\textbf{URL}: \url{https://w3id.org/iicongraph/}\\
\textbf{Documentation}: \url{https://w3id.org/iicongraph/docs/}
\section{Introduction}

Using Linked Open Data (LOD) in the context of cultural heritage (CH) simplifies the organization, publication, connection, and reuse of knowledge within this domain, and also provides a structure capable of expressing the complex relationships that can emerge between CH artifacts \cite{Lodi2017}.
Over the years, numerous Knowledge Graphs (KGs) have emerged that contain triples on cultural heritage, including those referenced in \cite{arco,wikidata,europeana,dbpedia,artgraph}. While some serve a general purpose and deal with various domains, others have been specifically crafted to incorporate and represent information about cultural heritage.
However, recent studies \cite{isdcsubjectenough} demonstrate that, in the artistic domains of iconography and iconology\footnote{Iconography is the study and interpretation of visual symbols and images, often within the context of art or visual representation. It involves the identification and analysis of symbols, motifs, and elements within images in an artwork. On the other hand, iconology involves the interpretation of images in a broader cultural and historical context, exploring the deeper layers of meaning, cultural ideologies, and socio-political influences associated with visual representations.\cite{panofsky_studies_1972}}, current KGs show two main issues: (i) iconographic and iconological statements lack granularity or are dumped\footnote{Considering dumping as the phenomenon in which ''important information for which no appropriate field was found, was forced as plain text inside a descriptive field, easy for humans to read but forever lost to any automatic tool" \cite{dumping}} in free text descriptions \cite{sartini2021towards}, and (ii) cultural symbolism\footnote{Intended as the set of symbolic meanings that some cultural heritage objects (or the elements depicted in them) convey from specific cultural perspectives} is severely underrepresented. 

This paper addresses these gaps by presenting \textit{IICONGRAPH}, a KG developed from the iconographic and iconological statements of Wikidata \cite{wikidata} and ArCo \cite{arco}, first re-engineered following the ICON ontology \cite{iconpaper} structure, and then enriched with LOD on cultural symbolism taken from HyperReal \cite{marriageisapeach}. These two KGs were chosen because they showed the greatest potential in the evaluation work by Baroncini et al. \cite{isdcsubjectenough}. In that study, ArCo and Wikidata obtained the highest scores for the correctness of their iconographic and iconological statements, while showing limits when it comes to the level of granularity of these statements.\footnote{In the mentioned study, both KGs performed poorly on the "structure" evaluation, which dealt with the possibility to differentiate between iconographic, iconological, and symbolic subjects, among other criteria that will be explained in section \ref{sec:qev}} 
First, this same evaluation is conducted on IICONGRAPH to demonstrate its superior performance compared to the original sources, highlighting its impact through quantitative assessments. Second, the research potential of IICONGRAPH is tested by attempting to address domain-specific research questions that remained unanswered with the original data from Wikidata and ArCo.

The rest of the paper is divided as follows. Section \ref{sec:background} gives a background of the work by presenting the resources included in IICONGRAPH, the ontology behind it, and the technical and research-related limitations of the KGs that were chosen as the initial data sources. Section \ref{sec:development} follows by describing the development and release of IICONGRAPH. In section \ref{sec:qev} IICONGRAPH is evaluated using the same evaluation methodology proposed in \cite{isdcsubjectenough}. The following section \ref{sec:resev} describes how the re-engineered KG can now be used to answer domain specific research questions. Section \ref{sec:discussion} contains a discussion reflecting on the results of the quantitative and research based evaluations. Then, section \ref{sec:relwork} mentions related work about the generation of artistic knowledge graphs. Finally, section \ref{sec:conclusions} contains final reflections about this work and mentions possible future work.

\section{Background, problem statement, and research requirements}
\label{sec:background}

In this section, the resources used to develop and enrich IICONGRAPH are described. In the subsection about ArCo and Wikidata, their main issues in the representation of artistic interpretation domain are highlighted with examples. These highlighted constraints lead to the formulation of research questions, the answers to which are unattainable in the current versions of the KGs. Effectively addressing these research questions serves as a requirement and an evaluation criterion for IICONGRAPH.

\subsection{ICON ontology 2.0}

ICON \cite{iconpaper} is an ontology that conceptualizes artistic interpretations by formalizing the methodology of E. Panofsky \cite{panofsky_studies_1972}. According to Panofsky, when performing an artistic interpretation, the interpreter can consider three levels. At a pre-iconographic level, artistic motifs and their factual or expressional meanings are recognized both as single entities and as groups (or compositions). The recognition of a tree, of the action of running or the emotion of crying would be considered pre-iconographica. At an iconographical level, the same motifs are now recognized as what Panofsky calls images, and these images represent characters, symbols, personifications, specific places, events or objects (such as Rome, World War II, and Thor's Hammer). At the same level, the artwork can be seen as depicting a story or an allegory (Panofsky uses "Invenzione" as a general term conveying both stories and allegories). At an iconological level, the artworks are then analyzed in comparison with the cultural context in which they were created, and they become a vessel to convey more in-depth cultural meanings or representing cultural values or cultural phenomena. ICON was updated (in version 2.0) to include three shortcuts that directly link an artwork to the element of pre-iconographic, iconographic or iconologic levels that it depicts or represents \cite{icon20}. For a more comprehensive overview of ICON, refer to the documentation of IICONGRAPH\footnote{\url{https://w3id.org/iicongraph/docs/}} or to previous publications on the ontology \cite{icon20,iconpaper}.

\subsection{HyperReal}

\begin{figure}
    \centering
    \includegraphics[width=0.66\linewidth]{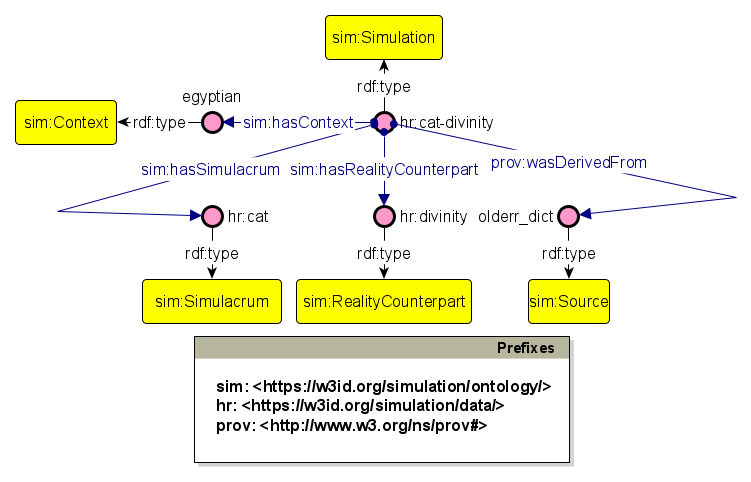}
    \caption{Graphical Rendering of the \textit{cat-divinity} simulation following the Simulation Ontology schema. The Simulacrum is the specific term used to represent symbols, while Reality Counterpart is the term used to represent symbolic meanings.}
    \label{fig:catdiv}
\end{figure}

HyperReal is a knowledge graph that contains more than 40,000 instances of symbolism, also called simulations. A simulation is a connection between a symbol (like a cat) and its symbolic meaning (such as divinity) in a particular cultural context (e.g. Egyptian). HyperReal information comes from various sources such as symbols dictionaries \cite{olderr_symbolism_2012}, and encyclopedias \cite{otto_mythological_2014} and is structured according to the Simulation Ontology framework. Figure \ref{fig:catdiv} shows the graphical rendering of the \textit{cat-divinity} simulation. The KG is available through its data dump at \url{https://w3id.org/simulation/data}. HyperReal data are aligned with the corresponding Wordnet \cite{fellbaum_wordnet_1998} and Babelnet \cite{navigli2010babelnet} synsets to facilitate the process of aligning external data with its symbols and symbolic meanings \cite{connectingsartini2023}.
HyperReal has been used in the back end of cultural heritage-related applications \cite{multivocal}, and as a data source for quantitative comparative cultural studies \cite{adhosymbol}. In the context of IICONGRAPH, it is used to enrich the potential symbolism of artworks. From the example above, a painting depicting a cat could be interpreted from an Egyptian point of view as symbolizing divinity. This kind of inference is agnostic from the intention of the creator of the work of art, but contributes to its understanding from a polyvocal and multicultural perspective.

\subsection{Wikidata}
 Wikidata \cite{wikidata} is a user-generated, open, comprehensive knowledge base, launched in 2012 by the Wikimedia Foundation, with a wide selection of content, available in various levels of detail and formats. It provides a platform for collaboration, sharing, integration, and a technology system for creating linked data. In the Digital Humanities (DH) domain, it is often used to annotate and improve project components, curating metadata to refine the interoperability of authority and local data sets about cultural heritage \cite{10.1093/llc/fqac083}. In the context of this work, the main focus was on the subset of Wikidata's statements regarding artworks and their depictions (the extraction methodology is explained in section \ref{sec:development}).

\subsubsection{Wikidata Analysis and Problem statement}
When analysing the limitations of Wikidata's iconographic and iconological statements, the main focus is on the property \textit{depicts} (wdt:P180) and its qualifiers. This property links an artwork with an element depicted in it. Its qualifiers, such as \textit{wears} (wdt:P3828), and \textit{expression, gesture, or body pose} (wdt: P6022), give more context to the depicted element. On the one hand, Wikidata contains more than 372,000 \textit{depicts} statements\footnote{Query last run in December 2023: \url{https://qlever.cs.uni-freiburg.de/wikidata/yKhv77}} when the subject is a painting (wd:Q3305213), which is a great starting point for digital art history studies. On the other hand, this property is used for all three levels of interpretations, flattening the expressivity of those statements. When it comes to expressing symbolism, there are some exceptions. For, instance \textit{symbolizes} (wdt:P4878) is a qualifier of \textit{depicts} that links a depicted element to what it symbolizes. However, the property is rarely used (only 63 statements related to paintings).\footnote{Query last run in December 2023: \url{https://w.wiki/8QyF}} \textbf{Therefore, the main issue with Wikidata is that when the data are present, the schema is lacking, and when the schema is present, the data is lacking.}

\subsubsection{Formulation of research questions for Wikidata}
Following the previous statements, in Wikidata it is not possible to retrieve what the most symbolic paintings are and how many serendipitous symbolic connections exist between paintings. Serendipitous connections are defined here as \begin{quote}
    "all the new connections that emerged between artworks in Wikidata, caused by the shared symbolic meaning \textbf{only}. The last part of the definition is essential because two painting may share a symbolic meaning [if] they share the symbol that symbolises it, while here the emphasis is on artworks that share the symbolic meaning that is conveyed by different symbols. For example, if Painting \textit{A} and Painting \textit{B} both depict a heart, they will share the potential symbolism of love because they share the same symbol, this would not be a serendipitous connections. Contrarily, if Painting \textit{A} contains a heart and painting \textit{C} contains a red rose, they share the symbolic meaning of love without sharing the same symbol, which leads to a serendipitous discovery. The concept of serendipity is to be taken in a broader sense, as the serendipitous connections are hidden at first, and are highlighted only by revealing the symbolic meaning of the depicted entities."\cite{connectingsartini2023}
\end{quote}
By using the current data in Wikidata, zero serendipitous connections emerge.\footnote{Query last run in December: \url{https://w.wiki/6BZR}}
At the same time, it is also currently challenging to distinguish between the pre-iconographical and the iconographical elements depicted in Wikidata's painting, a task that becomes even more difficult if the objective is distinguishing between the specific types of iconographical subjects (characters, places, attributes, etc...)

Given these premises, the following research questions have been formulated and will be answered in IICONGRAPH.

\begin{itemize}
    \item [RQ1.1]To what extent can linked open data be leveraged to expose serendipitous symbolic connections in Wikidata?
    \item [RQ1.2] Which artworks emerge as being the most symbolic?
    \item [RQ2.1] How are pre-iconographic and iconographic depictions distributed across Wikidata's \textit{depicts} statements in paintings?
    \item [RQ2.2] Among iconographical elements, which are the main classes (characters, places, attributes) that emerge as the most frequent?
\end{itemize}

The first research hypothesis linked to RQ1.1 and RQ1.2 is that after enriching Wikidata with HyperReal, the number of serendipitous connections will substantially increase, and after that it will be possible to rank Wikidata's painting according to their \textit{symbolic temperature}. Addressing RQ2.1 and RQ2.2, by re-engineering the statements in Wikidata according to ICON ontology, it will be possible to distinguish and measure the distribution of pre-iconographic and iconographic elements.

\subsection{ArCo}
ArCo \cite{arco} is a knowledge graph that describes a wide spectrum of artifacts from Italian cultural heritage, containing items belonging to architectural, ethnographic, and artistic domains. It follows the structure of the ArCo ontology, spread into different modules to address different levels of description of cultural heritage. In the context of this work, only a subset of statements related to artworks (which belong to the class \textit{ HistoricalOrArtisticProperty}) will be considered, with more limitations that will be explained in the following subsection.

\subsubsection{ArCo Analysis and Problem statement}
ArCo was created by applying Natural Language Processing algorithms to the OCR (Optical Character Recognition) version of printed catalogs. Consequently, even if some of the more technical information was converted into URIs and single nodes in the KG, a great deal of free-text information remains, especially about subjective domains like iconographic readings. Therefore, most of the information regarding iconographical and iconological statements is dumped in a free-text description, not exploiting the full potential of linked open data. On the one hand, this puts ArCo in a worse starting position compared to Wikidata, which expresses almost all of the information through URIs and limits the free-text fields. On the other hand, some of the descriptions in ArCo contained detailed interpretations about artworks, even separating pre-iconographic subjects from iconological meanings conveyed by artworks. Additionally, the descriptions have a very linear structure with repeating patterns, especially those related to a series of Italian billboards created in the 20th century. In the current version of ArCo, it is challenging to study the correlations between specific iconographic and pre-iconographic subjects and the cultural event/product they promote (iconological level).

\subsubsection{Formulation of research questions for ArCo}

Given that the starting point of ArCo is worse compared to Wikidata, only one set of research questions was formulated, namely:
\begin{itemize}
    \item [RQ3.1] What are the most common iconological meanings associated with Italian Billboards from the 20th century?
\end{itemize}
The research hypothesis is that by transforming the free-text description into structured data following ICON, it will be possible to isolate and then measure the frequency of iconological meanings. 

\section{IICONGRAPH Development and Release}
\label{sec:development}
This section describes how IICONGRAPH was developed and released. Different strategies were adopted for the development according to the issues mentioned in Section \ref{sec:background} for the two sources. The main distinction between the two sources is that while Wikidata provides information about the potential relationships between depicted entities (via the qualifiers), requiring a full description using the ICON ontology, ArCo's descriptions are very linear; therefore, only the shortcuts introduced in ICON 2.0 are necessary to describe such information.\footnote{As further proof for this decision, in the work that presents ICON 2.0 \cite{icon20}, ArCo's linear descriptions are used by the authors as one of the reasons to justify the need to create a simplified version ICON}

\subsection{Wikidata's Conversion}
\begin{figure}
    \includegraphics[width=0.5\linewidth]{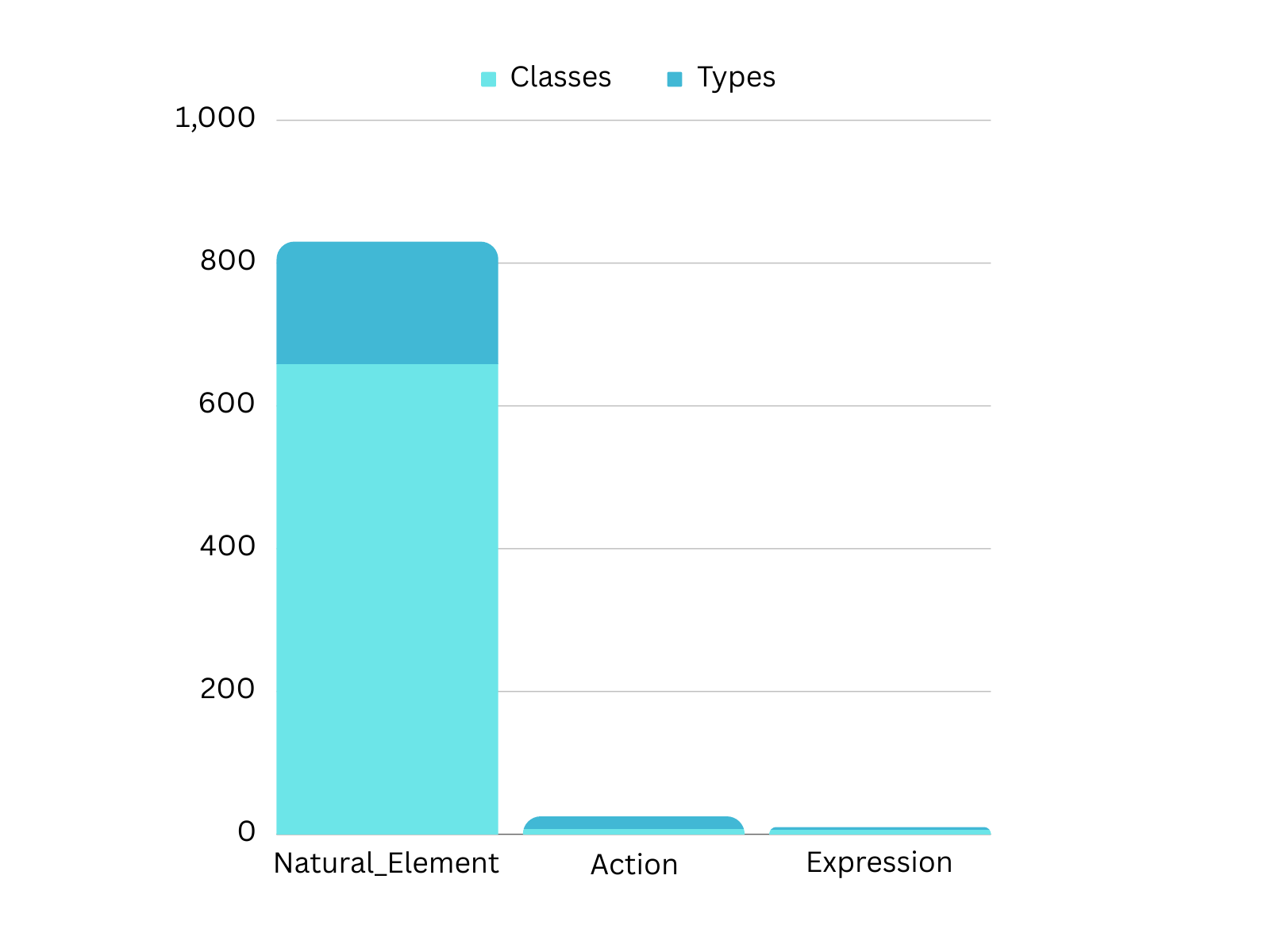}%
        \includegraphics[width=0.5\linewidth]{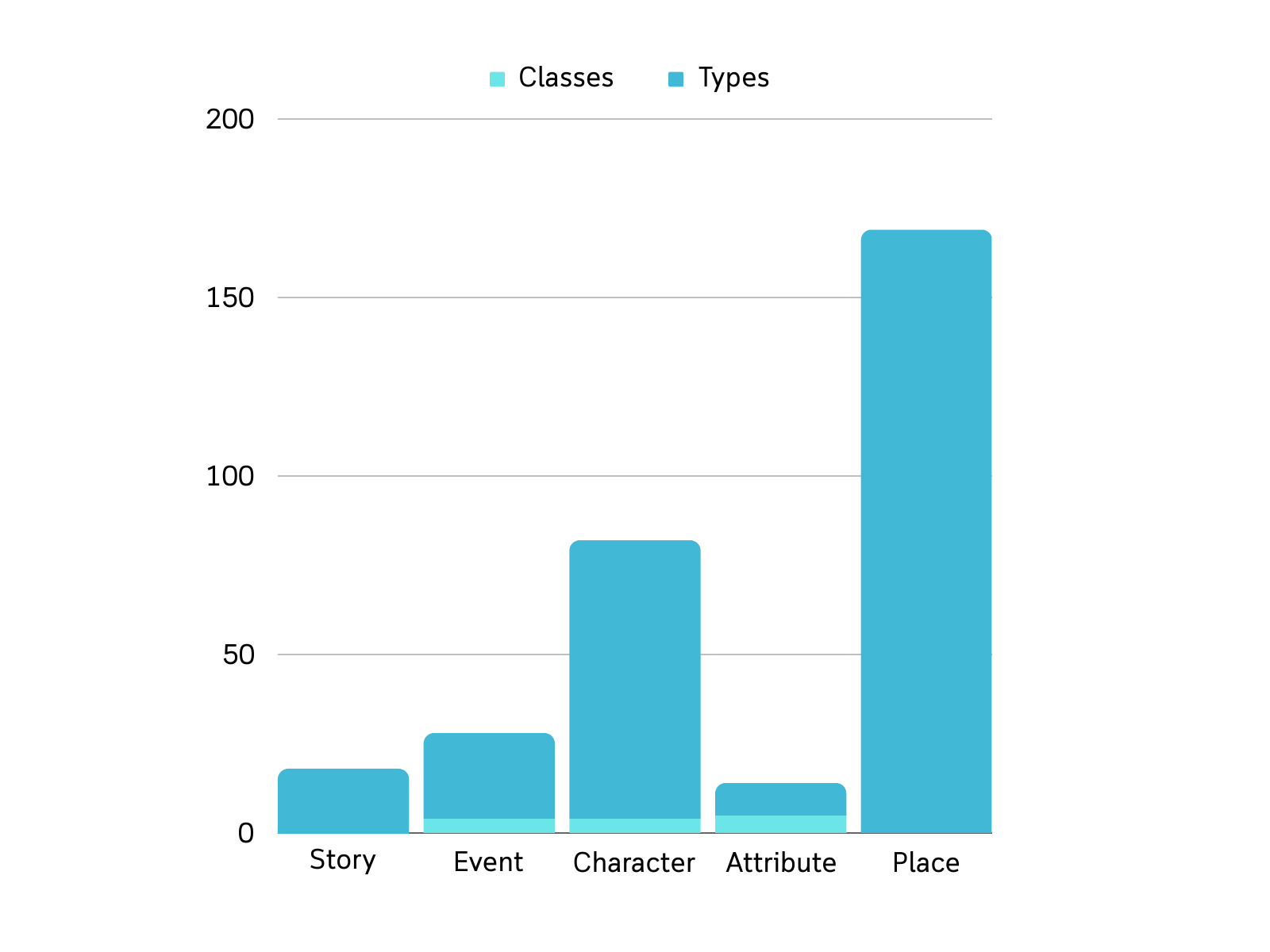}
            \caption{On the left: manual matching result between Wikidata's types and classes and ICON's classes related to pre-iconographical elements. On the right: manual matching result between Wikidata's types and classes and ICON's classes related to iconographical elements.}
                \label{fig:districon}
\end{figure}

The general pipeline adopted to convert Wikidata was (i) assigning the depicted entities to the classes of ICON, (ii) extracting data about paintings, (iii) aligning them to HyperReal, and then (iv) re-engineering the statements following ICON ontology.
To align Wikidata's depicted entities with ICON classes, we adopted a methodology involving the annotation of the depicted entity types and classes expressed in Wikidata through the properties \textit{instance of} (wdt:P31) and \textit{subclass of} (wdt:P279). Given the impracticality of manually annotating more than 60,000 individual depicted entities, I focused on annotating the top 700 classes and types, ordered by the number of depicted elements assigned to them. The top 700 covered more than 85\% of the total entities. To ensure objectivity, a no-ambiguity policy guided the single annotator. Each type or class was analyzed on Wikidata using a SPARQL query to verify that all related entities could match the designated ICON class; otherwise, the type or class was discarded. Figure \ref{fig:districon} illustrates the distribution of assigned classes and types for pre-iconographical and iconographical elements. After this alignment, the information about the paintings, their depicted elements, the types and classes of the depiction, and their qualifiers were extracted via a SPARQL query. A total of almost 150,000 paintings and their related metadata were extracted. To align Wikidata's entities with HyperReal's symbols for the enhancement, an alignment done in previous work was reused \cite{multivocal}. The conversion process occurred in a Python environment, using the RDFlib library\footnote{\url{https://rdflib.readthedocs.io/en/stable/}} The conversion of Wikidata yielded more than 29,000,000 triples. More than 3,000,000 symbolic interpretations were inferred, due to the alignment to HyperReal, with an average of around 20 interpretations per painting. For a more detailed description of Wikidata's conversion, refer to the documentation.

\subsection{ArCo's Conversion}
\begin{figure}
    \centering
    \includegraphics[width=0.99\linewidth]{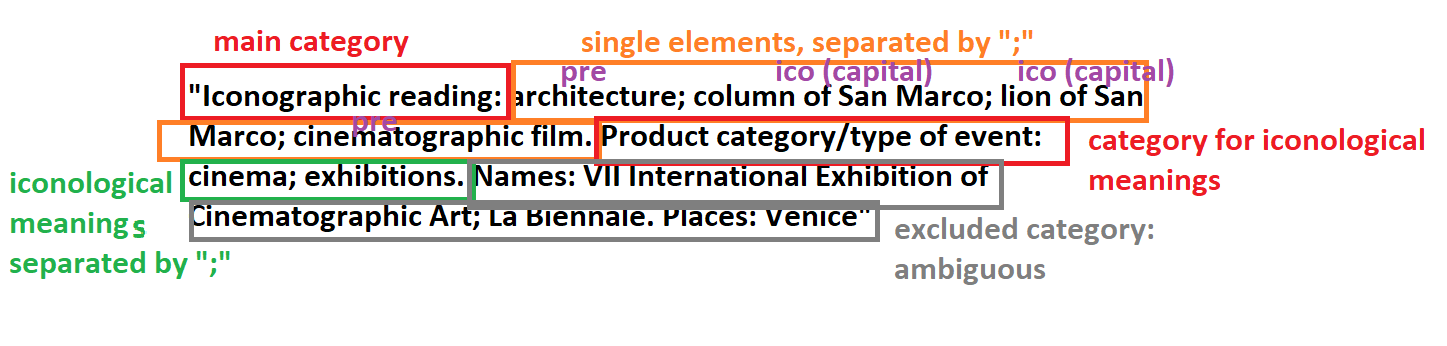}
    \caption{Visual Example of a standard description in ArCo and the parsing steps applied to it}
    \label{fig:parsarco}
\end{figure}

For the conversion of ArCo, only the shortcut version of ICON was necessary, eliminating the need to assign elements from free-text descriptions to individual ICON classes. Instead, the depicted elements were categorized into the macrogroups of pre-iconographical, iconographical, and iconological. The process involved extracting ArCo's data using a regex pattern to capture "Iconographic Reading:" (in Italian, "Lettura Iconografica:") in artwork descriptions linked via the Dublin Core description property (\texttt{dc:description}). Following the extraction of approximately 23,000 artworks and their descriptions (about 1\% of ArCo's total number of artworks), the structure of the descriptions was analyzed to identify other patterns that could be recognizable by an NLP algorithm for conversion. It was noticed that standard descriptions are organized into categories separated by a standard use of punctuation. All descriptions that did not meet this standard were discarded (around 3,000). All iconological meanings, in the context of billboards, were determined to be after the category "Product category/type of event" (in Italian, "Categoria Merceologica/tipo di evento"), where the promotional aspect was described. Ambiguous categories were excluded, and a straightforward approach was employed to distinguish between pre-iconographical and iconographical levels. If an element in the description was written with a capital letter, it was assigned to the iconographic level, otherwise to the pre-iconographical level. Figure \ref{fig:parsarco} visually shows the rationale behind the assignment and parsing of descriptions, exemplified by the artwork available at \url{https://w3id.org/arco/resource/HistoricOrArtisticProperty/0500659063}. Before conversion, all descriptions were translated into English using the Google Translate API. Given the simplicity of the texts, this translation did not generate evident errors. Single elements were linked to HyperReal through string matching.
The conversion of ArCo yielded 767,888 triples, which is significantly less than Wikidata because of the difference in number of artworks (150,000 against 20,000), and also because the simplified version of ICON is much less verbose.  A total of 457,747 automatic interpretations were generated due to the match with HyperReal.

\subsection{IICONGRAPH Release}
\label{sec:release}
IICONGRAPH was released according to the FAIR principles \cite{Wilkinson2016}.
The w3id service was used to obtain persistent URIs for the namespace \url{https://w3id.org/iicongraph/data/}, documentation \url{https://w3id.org/iicongraph/docs/} and analysis related to the research case studies \url{https://w3id.org/iicongraph/casestudies/}. The same information is accessible via the GitHub repository \url{https://github.com/br0ast/iicongraph/}. The prefix \textit{iig}, used in the KG, was registered in \url{http://prefix.cc}. The KG is stored in Zenodo, accessible via \url{https://zenodo.org/doi/10.5281/zenodo.10294588}. The metadata about the dataset and the provenance of the data is defined in a separate file \footnote{\url{https://github.com/br0ast/iicongraph/blob/main/data/IICONGRAPH\_catalogue.ttl}} following the DCAT standard.\footnote{\url{https://www.w3.org/TR/vocab-dcat-3/}} The ICON ontology used as its schema also respects the FAIR principles, obtaining a score of 90\% on the FOOPS tool \cite{foops}.

\section{Quantitative Evaluation}
\label{sec:qev}
In this section, IICONGRAPH is quantitatively evaluated following the methodology defined by Baroncini et al. \cite{isdcsubjectenough}. Three versions of IICONGRAPH will be evaluated, namely IICONGRAParco, which contains only the re-engineered statements from ArCo, IICONGRAPHwikidata, which contains only the re-engineered statements from Wikidata, and IICONGRAPHglobal, containing all the triples.
The assessment method considers six criteria, divided into two macro-areas: content and structure. Content considers the evaluation of the correctness of artistic interpretation statements (CR1), and the evaluation on the completeness of artworks interpretations (CR2) (i.e., whether the interpretation mentions, when needed, pre-iconographical, iconographical, and iconological statements). Then, structure addresses the richness of the schema describing the artworks (CR3), the entity linking of artworks with external sources is measured by CR4, CR5 measures how the URIs of the depicted subjects are linked within the same dataset (in technical terms, the outdegree of the subjects' nodes in a graph) and CR6 measures the number of references to external taxonomies of art and culture. All categories are given a weight (CR1,2,3 have a weight of 1, CR4,5 have a weight of 0.6 and CR6 has a weight of 0.8) and the possible scores for each go from 0 to 1. Given the reengineering tasks performed to create IICONGRAPH, this work could only influence the criteria CR2, CR3, CR5 as it does not deal with changing wrong interpretations (CR1), linking artworks between different datasets (CR4), or referring to external taxonomies of art and culture (CR6).

Following the methodology presented in \cite{isdcsubjectenough}, CR2 was calculated by averaging the scores of two annotators that evaluated the description of 100 artworks. The annotators had to decide how many interpretation levels they expected for the artwork. The general guidelines of \cite{isdcsubjectenough} say that artworks depicting a landscape usually are only interpreted via a pre-iconographical level, most portraits have both pre-iconographical and iconographical meanings, and allegorical, religious, and culturally relevant scenes (depiction of wars, special events for a country or culture) can usually be described using all levels. After averaging the evaluation, IICONGRAPHglobal obtained 0.92, IICONGRAPHarco 0.958 and IICONGRAPHwikidata 0.97. CR3 was calculated through a comparison of the ICON ontology structure with the golden standard in \cite{isdcsubjectenough}\footnote{In summary, a perfect schema would be able to describe actions, preiconographical elements, stories, allegories, iconographical subjects, symbols, iconological subjects, cultural phenomena, and should be able to be used in combination with a taxonomy or controlled vocabulary of art and culture}. Given that the schema behind IICONGRAPH is the ICON ontology, developed to describe all the information mentioned in the golden standard, the score of all the versions of the KG in this category was set at 1. CR5 was computed via SPARQL queries on the data, first counting how many subjects in the data were linked to at least more than 1 artwork, and then dividing this number by the total number of subject recognized. The scores obtained are 0.5771 for IICONGRAPHarco, 0.4573 for IICONGRAPHwikidata and 0.4337 for IICONGRAPHglobal. Since CR1, CR4, and CR6 were not affected by the changes, IICONGRAPHwikidata and IICONGRAPHarco maintain their scores of \cite{isdcsubjectenough}, while IICONGRAPHglobal receives an average of the two scores.
Table \ref{tab:quant} shows the scores compared to the other datasets analyzed in \cite{isdcsubjectenough}. In Section \ref{sec:discussion} the results are analyzed and discussed. All scripts and queries related to the quantitative evaluation are available in the documentation at \url{https://w3id.org/iicongraph/docs/}.
\begin{table}[]
\resizebox{.99\columnwidth}{!}{%
\begin{tabular}{@{}lllllllllllll@{}}
\toprule
 & \multicolumn{1}{l|}{CR1 UF} & \multicolumn{1}{l|}{CR2} & \multicolumn{1}{l|}{CR3} & \multicolumn{1}{l|}{CR4 UF} & \multicolumn{1}{l|}{CR5} & \multicolumn{1}{l|}{CR6 UF} & \multicolumn{1}{l|}{Content} & \multicolumn{1}{l|}{Rank Cont} & \multicolumn{1}{l|}{Structure} & \multicolumn{1}{l|}{Rank Structure} & \multicolumn{1}{l|}{Overall} & Rank overall \\ \midrule
IICONGRAPHglobal & 0.9023 & \textbf{0.92} & \textbf{1} & 0.3508 & 0.4737 & 0.1404 & \textbf{0.9111} & \textbf{2} & \textbf{0.5357} & \textbf{2} & \textbf{0.7234} & \textbf{2} \\
IICONGRAPHarco & 0.8278 & \textbf{0.958} & \textbf{1} & 0.0026 & \textbf{0.5771} & 0.1238 & \textbf{0.8929} & \textbf{3} & \textbf{0.4823} & \textbf{3} & \textbf{0.6876} & \textbf{3} \\
IICONGRAPHwiki & 0.9768 & \textbf{0.97} & \textbf{1} & 0.699 & 0.4573 & 0.157 & \textbf{0.9734} & \textbf{1} & \textbf{0.6065} & \textbf{1} & \textbf{0.7899} & \textbf{1} \\ \midrule
ArCo & 0.8278 & 0.74 & 0.3333 & 0.0026 & 0.172 & 0.1238 & 0.7839 &  & 0.1790 &  & 0.4815 &  \\
Wikidata & 0.9768 & 0.74 & 0.6667 & 0.699 & 0.367 & 0.157 & 0.8584 &  & 0.4773 &  & 0.6678 &  \\ \midrule
Fondazione Zeri & 0.9925 & 0.5117 & 0.1111 & 0.0005 & 0.266 & 0.5449 & 0.7521 &  & 0.2356 &  & 0.4939 &  \\
Nomisma & 0.9768 & 0.5 & 0.2222 & 0 & \textbf{0.749} & 0.0001 & 0.7384 &  & 0.2239 &  & 0.4811 &  \\
SARI & 0.849 & 0.3783 & 0.1111 & 0.997 & 0.5 & 0 & 0.6136 &  & 0.3364 &  & 0.475 &  \\
Europeana & 0.4688 & 0.236 & 0.1111 & 0.0073 & \textbf{0.6122} & 1 & 0.3524 &  & 0.4276 &  & 0.39 &  \\
ND\_Hungary & 0.13 & 0.5392 & 0.1111 & 0 & 0 & 0 & 0.3346 &  & 0.037 &  & 0.1858 &  \\
DBpedia & 0.655 & 0.7242 & 0.2222 & 0.994 & 0.41 & 0 & 0.6896 &  & 0.3549 &  & 0.5222 &  \\
Yago & 0.99 & 0.4825 & 0.1111 & 1 & 0.1675 & 0 & 0.7362 &  & 0.2705 &  & 0.5034 &  \\ \bottomrule
\end{tabular} }
    \caption{Overall results of the quantitative evaluation applied to IICONGRAPHglobal, IICONGRAPHwikidata, and IICONGRAPHarco compared to the results in the state of the art performed in \cite{isdcsubjectenough}. UF labeled criteria signal that they were not affected by the changes. Ranking signals only the overall top 3 for each category.}
    \label{tab:quant}
\end{table}

\section{Research-based Evaluation}
\label{sec:resev}
This section shows how IICONGRAPH can be used to address the research questions formulated in Section \ref{sec:background}.
 Regarding RQ1.1 and RQ1.2, the methodology consisted of extracting the data on paintings and their symbolic depictions through a SPARQL query performed on IICONGRAPHwikidata. Around 79,000 paintings were associated with a symbolic meaning shared by more than one symbol. Artwork connections were computed using Python, with an iterative process comparing each depicted element between pairs of paintings. The calculation involved determining how many symbolic meanings were shared between the depicted elements of the pairs. At the end of the calculation, \textbf{2,481,489,938 serendipitous connections were exposed}. RQ1.2 was tackled with a SPARQL query, revealing the top 10 most symbolic paintings. \textbf{"Entrance into the Ark" (wd:Q209050) by Jan Brueghel The Elder} tops the list with almost 1,500 associated simulations; the rest of the top 10 are detailed in Table \ref{tab:top10}. In general, paintings with a multitude of animals and plants were associated with most symbolic meanings.

Similarly, simple SPARQL queries facilitated the examination of the distribution of pre-iconographic and iconographic representations in Wikidata (RQ2.1, RQ2.2), revealing that \textbf{ 64.86\% of the depicted elements belong to the pre-iconographical level}. Among iconographic elements, \textbf{Characters are the most recognized, with almost 100,000 occurrences}. The results of this analysis are presented in Table \ref{tab:icostatwiki}.

\begin{table}[ht]
\begin{tabular}{@{}llllll@{}}
\toprule
Level of   interpretation & Total & Unique & Specific Element & Total & Unique \\ \midrule
\textbf{Pre-iconographical} & \textbf{224,981} & \textbf{5,131} & \textbf{natural elements} & \textbf{220,463} & \textbf{4,938} \\
 &  &  & actions & 4,189 & 147 \\
 &  &  & expressions & 4,667 & 65 \\ \midrule
Iconographical & 121,893 & 37,667 & \textbf{characters} & \textbf{98,354} & \textbf{27,847} \\
 &  &  & events & 817 & 399 \\
 &  &  & stories & 3,436 & 3,436 \\
 &  &  & attributes & 791 & 51 \\
 &  &  & places & 17,050 & 5,438 \\ \bottomrule
\end{tabular}
\caption{Distribution of Pre-iconographical and Iconographical statements in Wikidata extracted from IICONGRAPHwikidata}
\label{tab:icostatwiki}
\end{table}

RQ3.1 followed a similar approach. Through a SPARQL query on IICONGRAPHarco, the number of paintings associated with each iconological meaning was determined. The top 10 iconological meanings are presented in Table \ref{tab:top10arco}. In particular, \textbf{ the iconological meaning most referred to in 20th century billboards is the promotion of tourism}.

\begin{table}[ht]
\begin{tabular}{@{}ll@{}}
\toprule
Iconological Meaning & Painting \# \\ \midrule
\textbf{iig:promotionOfTourism} & \textbf{4,572} \\
\textbf{iig:promotionOfExhibitions} & \textbf{3,604} \\
\textbf{iig:promotionOfTourismPromotionBodies} & \textbf{3,380} \\
iig:promotionOfInformationAndCommunication & 2,932 \\
iig:promotionOfFoodIndustry & 2,814 \\
iig:promotionOfCulturalEvents & 2,219 \\
iig:promotionOfTransport & 2,211 \\
iig:promotionOfSport & 1,928 \\
iig:promotionOfTrade & 1,911 \\
iig:promotionOfAgriculture & 1,694 \\ \bottomrule
\end{tabular}
\caption{Top 10 iconological meanings associated with the most artworks in ArCo}
\label{tab:top10arco}
\end{table}

In summary, post-reengineering and enrichment, all research questions formulated in Section \ref{sec:background} were effectively addressed. All scripts and queries developed to address these research questions are provided at \url{https://w3id.org/iicongraph/casestudies} to ensure the transparency and reproducibility of the results.

% \usepackage{booktabs}

\begin{table}[]
\begin{tabular}{@{}lll@{}}
\toprule
Painting ID & Painting Label & Sim\# \\ \midrule
\textbf{wd:Q66107722} & \textbf{Entrance into the   Ark} & \textbf{1,488} \\
\textbf{wd:Q18809786} & \textbf{Entry into Noah's   Ark} & \textbf{998} \\
\textbf{wd:Q321303} & \textbf{The Garden of   Earthly Delights} & \textbf{851} \\
wd:Q27980267 & Unknown Title & 758 \\
wd:Q2510869 & Concert in the   Egg & 747 \\
wd:Q463392 & Paradiesgärtlein & 723 \\
wd:Q20170089 & The Ark & 723 \\
wd:Q18917077 & The Garden of   Eden and the Creation of Eve & 721 \\
wd:Q29656879 & Earth or The   Earthly Paradise & 706 \\
wd:Q18573212 & The Animals   Entering Noah's Ark & 662
\end{tabular}
\caption{Top 10 of the most symbolic paintings in Wikidata, retrieved by a SPARQL query performed on IICONGRAPH. \textit{wd:} is the prefix for https://www.wikidata.org/wiki/entity/}
\label{tab:top10}
\end{table}

\section{Discussion of the results}
\label{sec:discussion}

After a thorough quantitative evaluation, the performance of IICONGRAPHglobal, along with its subsets IICONGRAPHwikidata and IICONGRAPHarco, outperforms the rest of the knowledge graphs examined in \cite{isdcsubjectenough} in both structure and content scores. The effectiveness of the re-engineering process is evident in the significant improvements observed, particularly in CR5 (subject intralinking potential) for ArCo, where it experienced an impressive increase (more than 300\%) from 0.172 to 0.5771. This increase is due to the generation of more subjects expressed in URIs, which increases the number of connections between artworks that share the same subject (now defereanceable compared to the previous text-only version).

The best-performing KG overall is IICONGRAPHwikidata, similarly to when the standard version of Wikidata was the top performer before the re-engineering process. Notably, despite the enhancements, ArCo still falls short in the structure criteria, with an overall structure score of less than 0.5. This limitation is attributed to issues such as references to external taxonomies and the alignment challenges between its artworks and those present in other knowledge graphs.

When it comes to the evaluation of the research, the RQs formulated were designed to reveal the new potential of IICONGRAPH, although these analyses are considered preliminary and show some conversion limitations. In fact, the automatic symbolic interpretations of artwork from a polyvocal point of view (given by HyperReal) could be the starting point for more in-depth analysis for art historians, as they only represent potential, creator-agnostic symbolic meanings. Regarding limitations, table \ref{tab:top10arco} displays elements that could be merged after performing entity disambiguation (i.e., promotion of tourism and promotion of tourism promotion bodies). Despite this, the results underscore the considerable advancement represented by IICONGRAPH and its subgraphs. They not only outperform the state-of-the-art quantitatively but also demonstrate their utility in addressing research questions that were previously unattainable using the original versions of the Knowledge Graphs. The improvements in both quantitative metrics and research potential underscore the significance of the re-engineering efforts and the enriched representations provided by IICONGRAPH.

\section{Related Work}
\label{sec:relwork}
This section provides an overview of the development of artistic knowledge graphs or related resources, highlighting the main differences from the work presented in this paper.

Artgraph \cite{artgraph} is a knowledge graph developed by combining data from DBpedia and Wikiart, including over 250,000 artworks and associated artists. Its objective is to integrate visual embeddings and graph embeddings from the knowledge graph for automated art analysis. Unlike IICONGRAPH, which focuses on art interpretation, Artgraph pursues a broader goal of predicting genres and styles via embeddings. However, an examination of Artgraph's properties reveals the same issues found in Wikidata and ArCo, such as the lack of granularity of iconographic and iconological statements due to the absence of interpretative depth. The connection between artworks and subjects relies on a generic "tag" property lacking specificity. Furthermore, the dataset does not incorporate symbolic representation.

ICONdataset \cite{baroncini_exploring_2023} is a manually annotated knowledge graph, containing more than 5,500 art historians' interpretations about more than 400 artworks. It shares with IICONGRAPH the adoption of the ICON ontology as its primary schema, and the aim of addressing art history research questions through quantitative methods rooted in LOD. While manual annotation, as employed by ICONdataset, affords complete supervision over the data, ensuring a high degree of accuracy, the inherent drawback lies in its time-consuming nature, evident in the relatively low numbers of artworks and interpretations. In contrast, IICONGRAPH adopts a semi-automatic approach, resulting in a significant disparity in both the quantity of artworks and interpretations between the two KGs. This distinction emphasizes the scalability and efficiency afforded by a semi-automatic process. Furthermore, IICONGRAPH's incorporation of the HyperReal enrichment introduces an additional layer of symbolic data, augmenting its comprehensiveness, reach, and potentialities in comparison to manually annotated counterparts like the ICONdataset.

MythLOD \cite{Pasqual_Tomasi_2022} is an LOD catalog that contains interpretations of more than 4,000 mythological works. It was created by converting a CSV manually populated by domain experts. Its main purpose is to represent in LOD both the methodology and rationale of the interpretations (iconographic, hermeneutic) and the bibliographic sources which supported the interpretations. However, when it comes to describing the main objects of the interpretations, it relies on the standard Dublin Core \footnote{\url{https://www.dublincore.org/specifications/dublin-core/dcmi-terms/}} subject property (\texttt{dc:subject}), which is extremely limited compared to the possibilities offered by the ICON ontology behind IICONGRAPH.

BACODI \cite{bruno2023odi} is a knowledge base developed to describe the relationship between the symbolic aspects of tarot cards and their narrative roles in the book "Il castello dei destini incrociati" by Italo Calvino. It uses the ODI ontology \footnote{\url{https://odi-documentation.github.io/materials/}} as its schema, developed specifically for the work. The knowledge base was created by converting a CSV manually annotated by a domain expert on Calvino's work, similar to ICONdataset and mythLOD. Despite including relationships that link the tarot card to their symbolism, it remains self-contained in Calvino's domain, not offering iconographic descriptions of the cards or symbolism that goes beyond the narrative roles in the book. In contrast, IICONGRAPH not only surpasses BACODI in quantitative scale but also provides a more expansive perspective on the iconographic and symbolic features within the artworks it contains.
Other datasets, such as \cite{europeana,zeri,dbpedia,yago,nomisma2} are not mentioned, as they are compared to IICONGRAPH through the evaluation in Section \ref{sec:qev}.
In summary, among the current artistic knowledge graphs, IICONGRAPH stands out for its dimension, detailed representation of iconographic and iconological statements, and enriched symbolic representation.

\section{Conclusion and future work}
\label{sec:conclusions}

This paper presented the development and evaluation of IICONGRAPH, a knowledge graph created by re-engineering the iconographic and iconological statements of ArCo and Wikidata. IICONGRAPH and its ArCo and Wikidata standalone versions outperformed the state-of-the-art of artistic KGs in both structure and content scores. Furthermore, the results of the qualitative evaluation based on research-based requirements demonstrate the suitability of IICONGRAPH to answer domain-specific artistic questions.
Future work is delineated into three main areas. First, the expansion of IICONGRAPH involves ingestion of additional statements from more artistic knowledge graphs, potentially leading to the development of a universal converter for turning artistic interpretation data into LOD following ICON ontology. Second, to enhance data accessibility, a web app or integration into existing platforms displaying ArCo's and Wikidata's information could be proposed. Third, in the realm of LLM, IICONGRAPH emerges as a valuable resource for developing question answering and chat-based systems focused on cultural heritage, addressing a gap identified in the literature regarding symbolic and iconographic knowledge \cite{garcia2020AQUA}. Additionally, the descriptions in ArCo and the RDF generated to create IICONGRAPH hold promise as fine-tuning arguments for an LLM capable of autonomously generating intricate iconographic linked open data from free-text descriptions via prompts.
In conclusion, IICONGRAPH stands as a robust and versatile resource that not only advances the understanding of artistic interpretation within the domains of art history and digital humanities, but also presents significant implications for the evolving landscape of Large Language Models, offering a promising avenue for further exploration and integration into the broader context of cultural heritage research.

%\begin{table}
%\caption{Table captions should be placed above the
%tables.}\label{tab1}
%\begin{tabular}{|l|l|l|}
%\hline
%Heading level &  Example & Font size and style\\
%\hline
%Title (centered) &  {\Large\bfseries Lecture Notes} & 14 %point, bold\\
%1st-level heading &  {\large\bfseries 1 Introduction} & 12 %point, bold\\
%2nd-level heading & {\bfseries 2.1 Printing Area} & 10 %point, bold\\
%3rd-level heading & {\bfseries Run-in Heading in Bold.} %Text follows & 10 point, bold\\
%4th-level heading & {\itshape Lowest Level Heading.} Text %follows & 10 point, italic\\
%\hline
%\end{tabular}
%\end{table}

%
% the environments 'definition', 'lemma', 'proposition', 'corollary',
% 'remark', and 'example' are defined in the LLNCS documentclass as well.
%

\bibliographystyle{splncs04}
\bibliography{bibliography}

\end{document}